\PassOptionsToPackage{table}{xcolor}
\documentclass[10pt,twocolumn,letterpaper]{article}

\usepackage[pagenumbers]{cvpr} % To force page numbers, e.g. for an arXiv version

\definecolor{cvprblue}{rgb}{0.21,0.49,0.74}
\usepackage[pagebackref,breaklinks,colorlinks,allcolors=cvprblue]{hyperref}

\newcommand{\second}[1]{\underline{#1}}
\usepackage{times,graphicx,amsmath,amssymb,booktabs,multirow,tabularx}
\usepackage[table]{xcolor}
\usepackage{tikz}
\usepackage{placeins}
\newcolumntype{Y}{>{\centering\arraybackslash}X}
\newcommand{\best}[1]{\textbf{#1}}
\def\confName{CVPR}
\def\confYear{2027}

\newcommand{\sg}{\operatorname{sg}}

\newcommand{\softmax}{\operatorname{softmax}}
\newcommand{\codec}{\mathcal{C}}

\definecolor{InteractTeal}{HTML}{008E89}
\definecolor{InteractNavy}{HTML}{172F46}
\newcommand{\interactvlalinks}{%
  \href{https://github.com/LingwuPro/InteractVLA}{%
    \raisebox{-0.15em}{\includegraphics[height=1.05em]{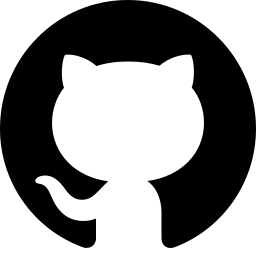}}%
    \hspace{0.35em}\textcolor{InteractNavy}{\textbf{Code}}}%
  \quad\textbar\quad
  \href{https://lingwupro.github.io/InteractVLA/}{%
    \raisebox{-0.22em}{\includegraphics[height=1.25em]{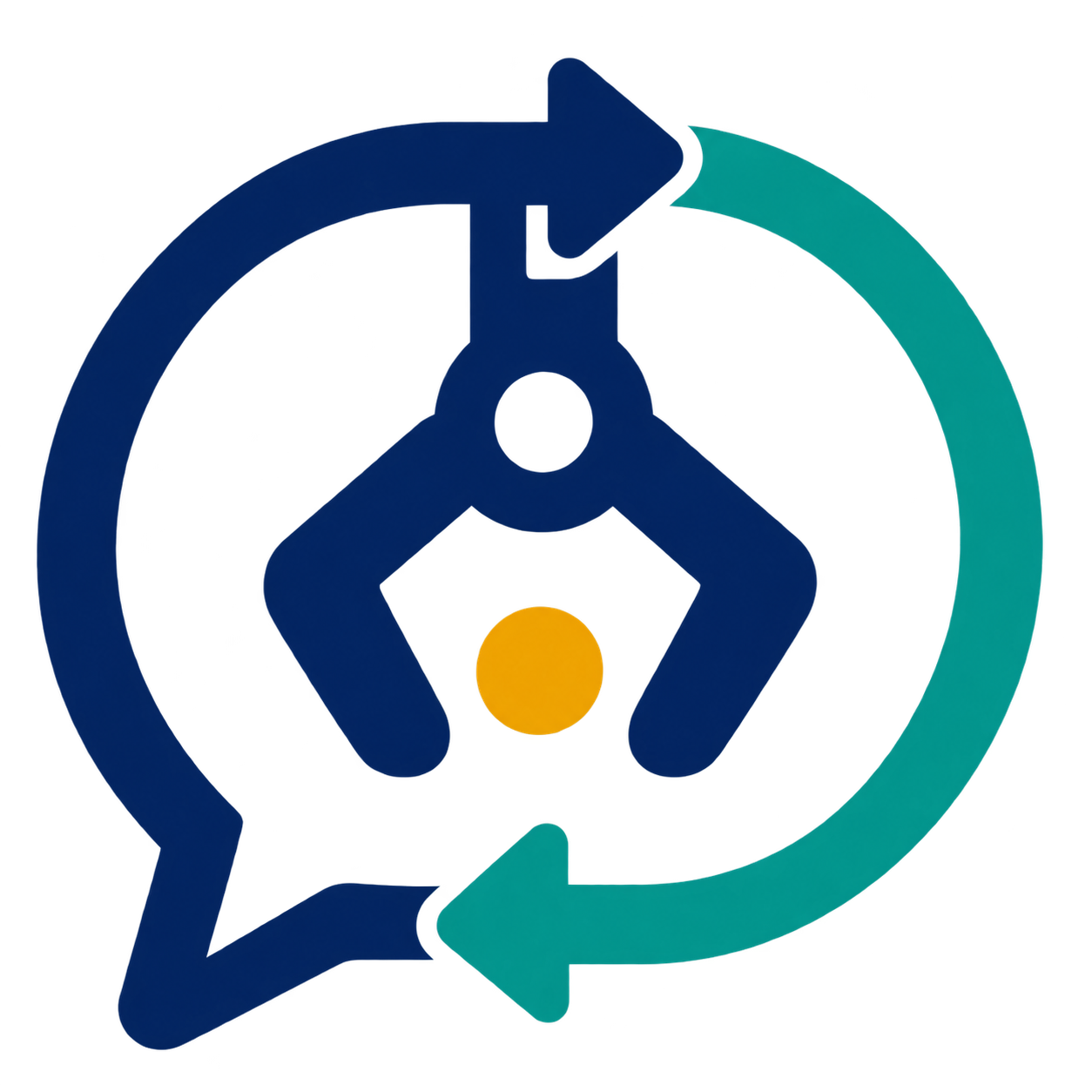}}%
    \hspace{0.35em}\textcolor{InteractTeal}{\textbf{Project Page}}}%
}

\title{\texorpdfstring{\raisebox{-0.08em}{\includegraphics[trim=9.3mm 2.15mm 0.7mm 2.15mm,clip,height=0.85em]{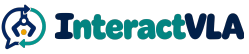}}}{InteractVLA}: Turning VLAs into Interactive Tools for Agentic\\ Robotic Manipulation}

\author{Chen Zhao\textsuperscript{\normalfont{1,2,3$\ast$}},
  Zipeng Ye\textsuperscript{\normalfont{3$\ast$}},
  Shifeng Bao\textsuperscript{\normalfont{1,2}},
  Yihan Lin\textsuperscript{\normalfont{1,2}},
  Haitao Shen\textsuperscript{\normalfont{1,2}},\\
  Yining Luo\textsuperscript{\normalfont{1}},
  Yuchen Xie\textsuperscript{\normalfont{3}},
  Jing Zhang\textsuperscript{\normalfont{1,2$\dagger$}} \\
  \textsuperscript{1} Renmin University of China, Beijing 100872, China\\
  \textsuperscript{2} Key Laboratory of Data Engineering and KnowledgeEngineering, MOE, China\\
  \textsuperscript{3} Meituan, Beijing 100102, China\\
  \texttt{<zhaochen100, zhang-jing>@ruc.edu.cn}
  \\[0.5em]
{\small\interactvlalinks}
}
\usepackage{graphicx}
\newcommand{\InteractVLAHeaderArt}{%
  \put(\dimexpr\paperwidth-10mm\relax,-5mm){%
    \raisebox{-\height}[0pt][0pt]{%
      \makebox[0pt][r]{\includegraphics[width=90mm]{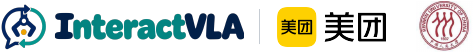}}%
    }%
  }%
}
\newcommand{\InteractVLAFirstPageHeader}{%
  \AddToHookNext{shipout/foreground}{\InteractVLAHeaderArt}%
}

\begin{document}
\InteractVLAFirstPageHeader
\maketitle
\vspace{-50px}
\begin{abstract}
Existing agentic manipulation systems treat pretrained vision-language-action (VLA) policies as opaque, language-only primitives: when a policy call fails under unfamiliar instructions or scene perturbations, the agent cannot steer its behavior and is forced into futile retries or costly manual takeover. We present InteractVLA, a framework that turns pretrained flow-based VLAs into geometrically steerable and inspectable tools through a bidirectional geometric interface with vision-language agents. In the forward direction, test-time geometric guidance translates an agent-specified end-effector target into input gradients through the action expert via a differentiable command-to-pose mapping, while soft-time target alignment treats arrival time as a latent execution index to match the target pose without distorting the policy's approach motion. In the reverse direction, action-informed agentic refinement exposes generated action chunks and calibrated camera reprojections back to the agent, allowing it to diagnose spatial discrepancies, revise targets for subsequent VLA calls, or locally adjust motion segments on top of the guided trajectory. Making the VLA controllable not only boosts online agentic tool use, but also unlocks effective recursive self-improvement (RSI), achieving state-of-the-art performance of 99.4\% on LIBERO-Plus, 48.62\% on RoboDojo, and 89.4\% on RoboCasa365.
\end{abstract}
\begingroup\def\thefootnote{$\ast$}\footnotetext{Equal contribution.}\endgroup
\begingroup\def\thefootnote{$\dagger$}\footnotetext{Corresponding author.}\endgroup
\begin{figure}
  \centering
  \includegraphics[width=0.5\textwidth]{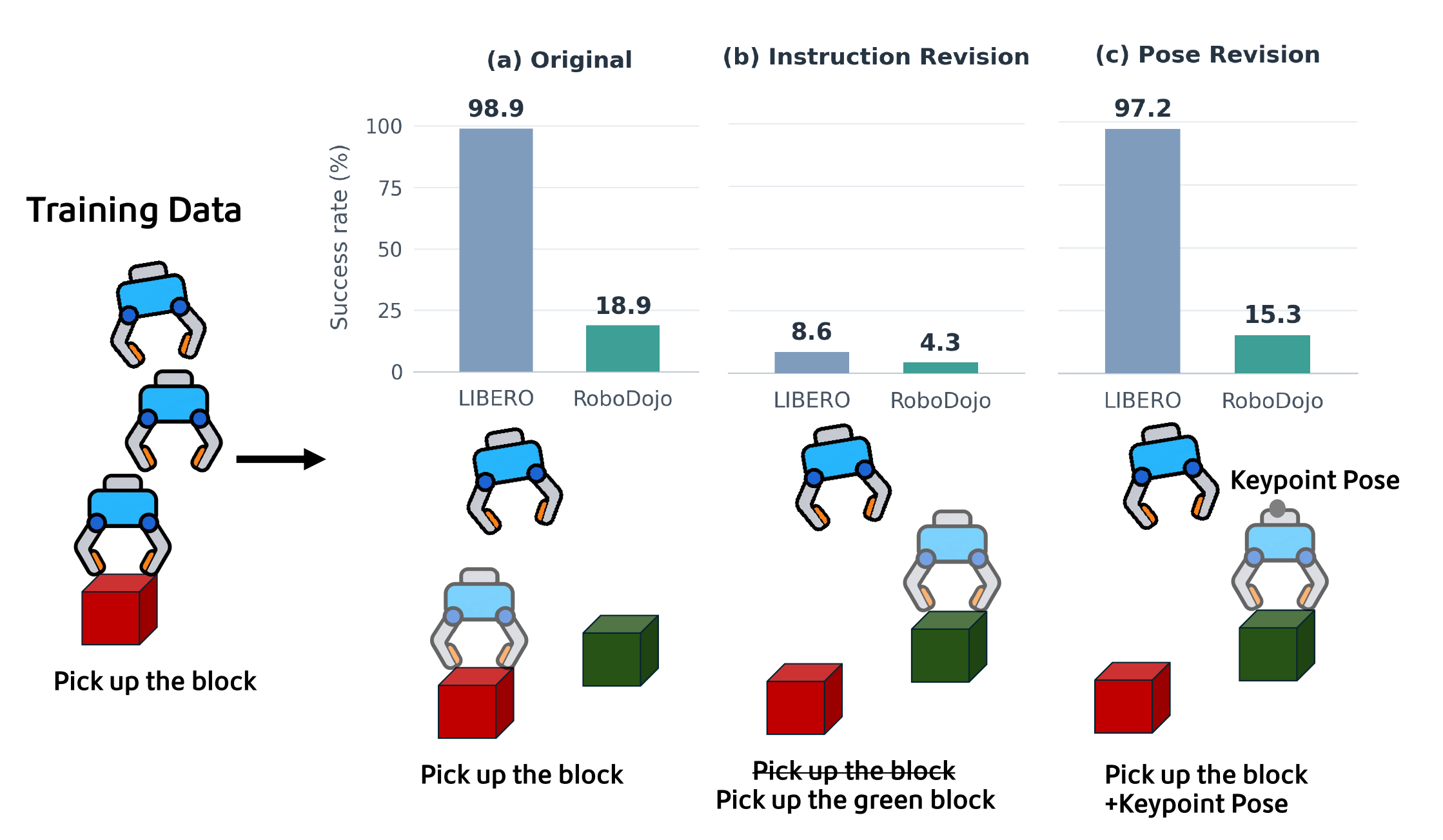}
  \caption{\textbf{Geometric guidance enables task-target generalization.} Changing the language instruction to specify a new target sharply reduces VLA success. Incorporating the target pose into action generation enables task transfer while maintaining high success rates.}
  \label{fig:img1}
\end{figure}
\begin{figure*}
  \centering
  \includegraphics[width=\textwidth]{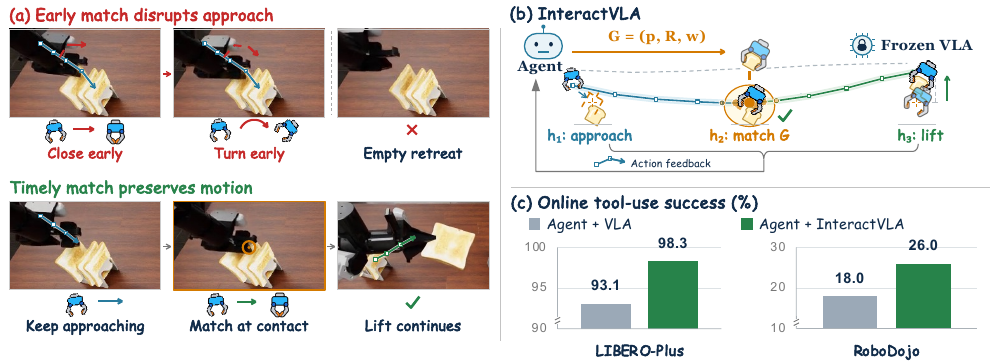}
  \vspace{-12px}
 \caption{\textbf{InteractVLA turns VLAs into interactive tools for agents.}
(a) Matching a sparse target too early disrupts the approach motion, causing premature gripper closure and rotation; timely alignment preserves the approach, contact, and lift sequence.
(b) InteractVLA combines test-time geometric guidance with soft-time target alignment to steer a frozen VLA toward agent-specified targets, while returning action feedback for target revision and local motion refinement.
(c) This interactive interface improves online tool-use success on both LIBERO-Plus and RoboDojo.}
  \label{fig:teaser}
\end{figure*}

\section{Introduction}
Combining vision-language agents with pretrained vision-language-action (VLA) policies promises an appealing division of labor for robotic manipulation: the agent handles long-horizon reasoning and task decomposition, while the VLA provides temporally coherent motor execution~\cite{harness,rvicl}. Yet in practice, existing VLAs hardly function as true interactive tools. Because standard VLAs only accept high-level language instructions and operate as opaque black boxes, the agent cannot steer how the policy executes a skill~\cite{zhou2025libero,li2026roborecover}. When a VLA call fails, simply commanding a retry is largely futile because it produces nearly identical motions under similar observations, whereas rephrasing the language instruction often pushes the VLA out of distribution and sharply degrades success (Figure~\ref{fig:img1}, middle). Across scenarios requiring generalization, the agent cannot rely on the VLA and must take over low-level control entirely, acting as a fallback supervisor that cleans up after an uncontrollable policy rather than a planner wielding a reliable tool~\cite{zhou2026improves,qu2026roboharness, choi2026vla,xu2026vla}.

This limitation stems from an inadequate interface between high-level reasoning and low-level control. Even when unfamiliar instructions or scene layouts prevent a VLA from grounding the right target, its pretrained spatial perception and motor priors, such as smooth approach trajectories, coordinated rotations, and grasp timing, remain highly capable~\cite{bai2025evolve,li2026simplevla}. As shown in Figure~\ref{fig:img1} (right), explicitly steering the end effector toward the intended interaction target allows the exact same policy to redirect its manipulation behavior and complete the task while preserving execution precision. Turning a VLA into a genuinely usable tool therefore requires moving beyond language-only invocation to a \emph{bidirectional geometric interface}: the agent must be able to specify explicit spatial targets to guide the VLA's action generation, and conversely inspect the resulting motion proposals to refine subsequent execution~\cite{pu2026tgm,chancan20253d,weng2026interaction}.

Building such a bidirectional interface at test time poses two coupled challenges. \emph{In the forward direction (Agent $\rightarrow$ VLA)}, the agent supplies a sparse target configuration to approach rather than a dense trajectory or exact arrival time. While the VLA should synthesize the intervening motion, naively applying a fixed-time penalty pulls the wrong execution step toward the target, whereas penalizing every step uniformly collapses the approach trajectory altogether. \emph{In the reverse direction (VLA $\rightarrow$ Agent)}, a binary success or failure signal is insufficient for recovery. Once guided by an agent-specified target, the VLA's generated action chunk already reflects the agent's spatial intent. By comparing the executed prefix and unexecuted continuation against visual observations, the agent can determine whether a failure requires calling the VLA again with a revised target or merely applying a lightweight local adjustment to the policy's well-structured trajectory, making recovery far more tractable than full manual takeover.

To address these challenges, we introduce \emph{InteractVLA}, a framework that turns pretrained flow-based VLAs into geometrically steerable and inspectable tools for agentic manipulation (Figure~\ref{fig:teaser}). Drawing on VLM-based policy steering~\cite{vls} and visual geometric constraints~\cite{voxposer,rekep}, InteractVLA bridges agent reasoning and action generation through three components. First, \emph{test-time geometric guidance} uses a differentiable command-to-pose mapping to translate an agent-specified end-effector target (position, orientation, and gripper state) into input gradients through the action expert within a single denoising trajectory, reshaping motion toward the agent's intent while returning control early if low confidence indicates an incompatible target. Second, \emph{soft-time target alignment} formulates arrival time as a latent execution index, jointly matching complete target poses at a shared arrival step that emerges naturally as the trajectory takes shape. Third, \emph{action-informed agentic refinement} exposes generated action chunks alongside calibrated camera reprojections and execution history to the agent, enabling it to diagnose spatial discrepancies, revise subsequent targets, or locally edit motion segments.

\begin{figure*}
  \centering
  \includegraphics[width=\textwidth]{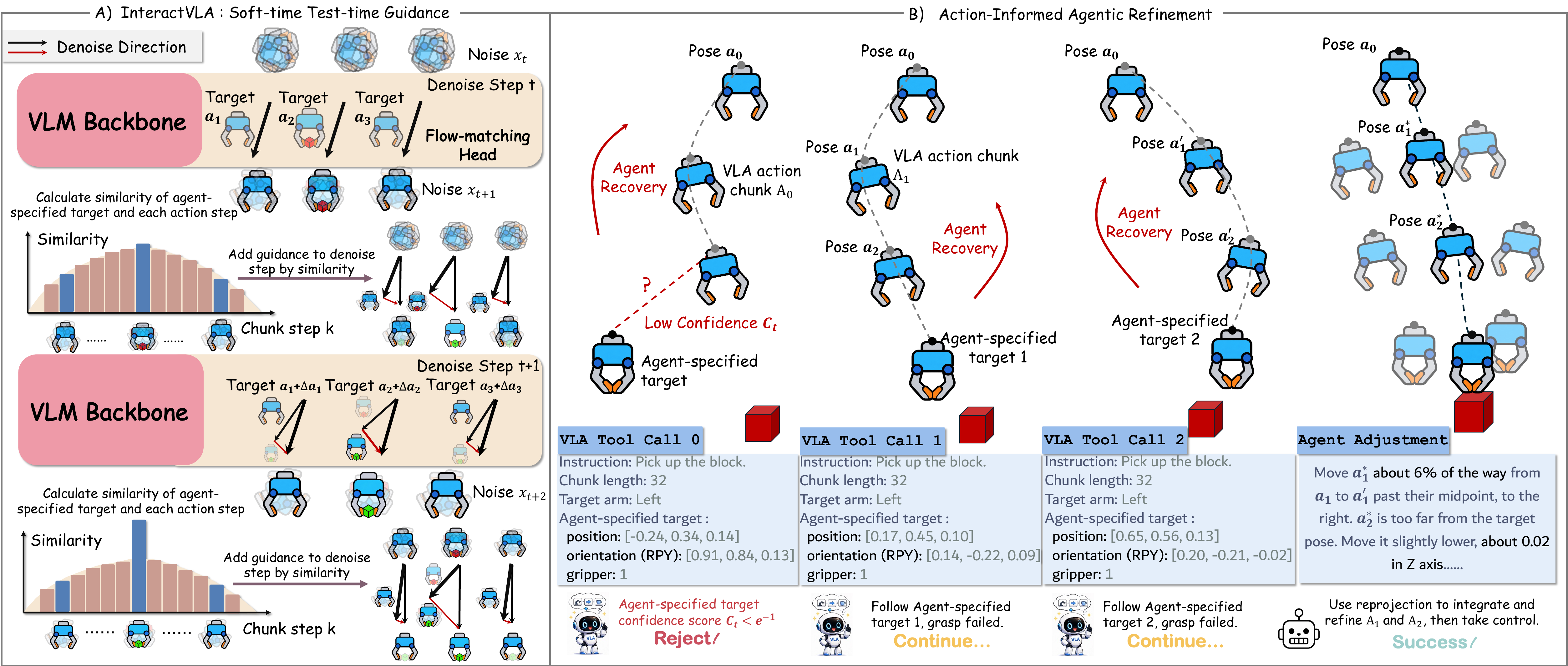}
  \caption{\textbf{InteractVLA combines geometric guidance with action-informed refinement.} \emph{Left:} At each denoising step, target similarity determines the guidance weights across the action chunk, progressively steering the generated motion toward the agent-specified target. \emph{Right:} Different targets elicit diverse motion proposals that the agent can reproject, compare, and locally refine to complete the task, while low-confidence targets trigger rejection and replanning.}
  \label{fig:main}
\end{figure*}
Because InteractVLA makes the VLA steerable rather than rigidly repeating a single action mode, it also provides a strong foundation for \emph{recursive self-improvement (RSI)}: across attempts, the agent iteratively sharpens how it specifies geometric targets and refines policy-generated motions. Across LIBERO-Plus~\cite{liberoplus}, RoboDojo~\cite{robodojo}, RoboCasa365~\cite{robocasa365}, and real-world manipulation, InteractVLA substantially outperforms standalone VLAs, language-only agent-VLA baselines, while reducing token consumption by over an order of magnitude compared with step-by-step agent-only control.

Our contributions are threefold:
\begin{itemize}
\item We introduce \emph{InteractVLA}, a framework that turns pretrained VLAs into interactive, controllable tools for agents via test-time geometric guidance, soft-time target alignment, and action-informed agentic refinement.
\item We show that making VLAs geometrically steerable and inspectable unlocks effective \emph{recursive self-improvement (RSI)}, allowing agents to progressively refine target specification and motion correction.
\item We achieve state-of-the-art performance across LIBERO-Plus (99.4\%), RoboDojo (48.62\%), and RoboCasa365 (89.4\%), with real-world evaluations confirming consistent gains in manipulation success and efficiency.
\end{itemize}
\section{Method}
\label{sec:method}
InteractVLA connects a vision-language agent and a pretrained flow-based VLA at test time through explicit geometric targets and motion feedback (Figure~\ref{fig:main}). Given observation $o_k$ and instruction $\ell$, the policy generates an action chunk $A_k\in\mathbb{R}^{H\times D}$ guided toward an agent-specified target $G_k=(p^*,R^*,w^*)$ comprising end-effector position, orientation, and gripper state. The agent then inspects the generated motion proposal to refine subsequent decisions.

\subsection{Test-Time Geometric Guidance}
\label{sec:ttg}
To steer a pretrained VLA toward the agent's geometric intent, we differentiate a pose objective through its action expert during denoising. Let $v_\theta(x_t,t;c_k)$ denote the flow velocity field conditioned on $c_k=(o_k,\ell)$, with generative time $t\in[0,1]$ running from noise to clean actions. From the one-step clean-action estimate
\begin{equation}
 m_t=x_t+(1-t)v_\theta(x_t,t;c_k),
 \label{eq:clean}
\end{equation}
a differentiable forward mapping $\codec_{o_k}$ converts native policy actions into command poses. At each execution step $h$, the complete-pose error $E_h$ aggregates normalized position, rotation, and gripper discrepancies relative to $G_k$ (summed across effectors at a shared index for bimanual control), directly linking the spatial target to native action commands.

Given a temporal objective $\mathcal{L}_t$ over step-wise pose errors, we steer the denoising trajectory via
\begin{align}
 g_t&=\nabla_{x_t}\mathcal{L}_t(\codec_{o_k}(m_t),G_k),\label{eq:grad}\\
 x_{t+\Delta t}&=\sg\!\left[x_t+\Delta t\bigl(v_\theta-\mathcal{B}_t(g_t)\bigr)\right],
 \label{eq:update}
\end{align}
where $\mathcal{B}_t$ scales the gradient, caps its magnitude relative to the native velocity $v_\theta$, and restricts guidance to an interior active window so endpoints retain the native flow field. The stop-gradient operator $\sg[\cdot]$ detaches successive integration states so each update differentiates solely through the current expert, whose input Jacobian couples corrections across the chunk.

To prevent incompatible targets from distorting generation, we monitor a target compatibility confidence $C_t$ comparing the minimum pose error $r_t=\min_h E_h$ and relative correction demand $d_t$ against tolerances $r_0$ and $d_0$:
\begin{equation}
 C_t=\exp\!\left[-\max\!\left(\frac{r_t}{r_0},\frac{d_t}{d_0}\right)\right].
 \label{eq:guidanceconfidence}
\end{equation}
If $C_t<e^{-1}$ after the first guided step, generation terminates early and returns control to the agent for replanning.

\begin{figure}
  \centering
  \includegraphics[width=0.45\textwidth]{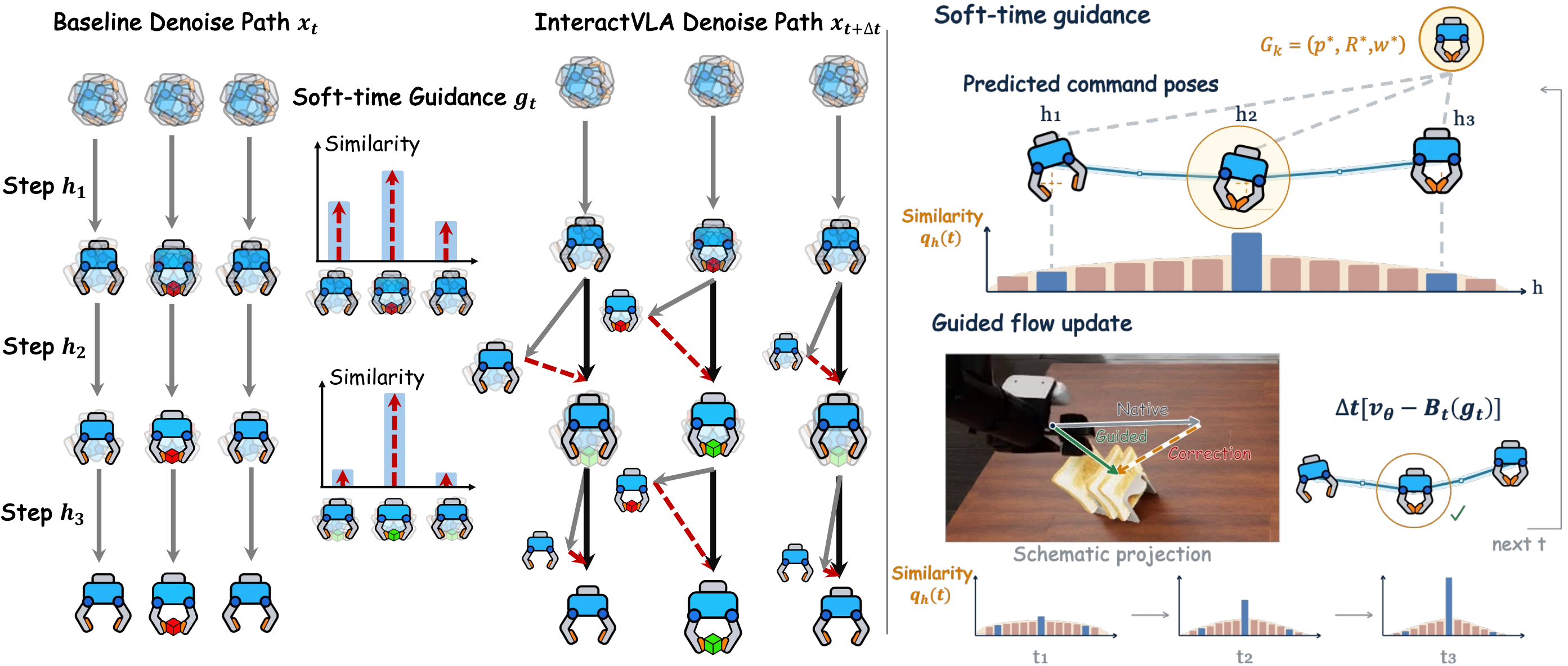}
  \caption{\textbf{Soft-time alignment progressively concentrates guidance on the best-matching action step.} As denoising proceeds, guidance increasingly focuses on the step most compatible with the agent-specified target, while the remaining steps retain their original motion structure so arrival time emerges naturally.}
  \label{fig:soft}
\end{figure}
\subsection{Soft-Time Target Alignment}
\label{sec:soft}
While the agent specifies a desired interaction pose, its appropriate arrival step within the horizon $H$ depends on task progress and the policy's speed. Imposing a fixed-step penalty forces an arbitrary timing assumption, whereas penalizing all steps uniformly collapses the approach motion toward the target. We instead treat arrival time as a latent execution index and softly align the target to the trajectory (Figure~\ref{fig:soft}).

Let $\tau=1-t$ and $\beta_t=t^2/\tau^2$. We aggregate the step-wise pose errors via a log-sum-exp objective:
\begin{equation}
 \mathcal{L}_{\rm soft}(t)
 =-\tau^2\log\!\left(\frac{1}{H}\sum_{h=1}^{H}
 \exp[-\beta_t E_h]\right).
 \label{eq:soft}
\end{equation}
Taking the gradient yields a temporal assignment $q_h(t)=\softmax_h(-\beta_t E_h)$ that distributes guidance across compatible arrival steps:
\begin{equation}
 \mathcal{L}_{\rm guide}(t)=\frac{\tau}{t}\mathcal{L}_{\rm soft}(t),
 \quad g_t=t\tau\sum_hq_h(t)\nabla_{x_t}E_h.
 \label{eq:fm}
\end{equation}
Because position, orientation, and gripper discrepancies enter $E_h$ jointly, they share the same temporal weight $q_h(t)$, ensuring the policy matches a complete interaction pose at a unified arrival step. As flow time $t$ advances and $\beta_t$ increases, $q_h(t)$ sharpens around the most compatible step while the multiplier $t\tau$ modulates overall guidance amplitude. This entropy-regularized temporal assignment lets the VLA jointly resolve the intervening approach trajectory and the target's exact arrival index.

\subsection{Action-Informed Agentic Refinement}
\label{sec:agent}
In the reverse direction, the guided action chunk serves as both an executable motion and an interpretable proposal whose approach vector, orientation profile, and grasp timing help diagnose spatial errors. Rather than returning a binary signal~\cite{chen2026show,jia2026agent}, InteractVLA exposes the full chunk, its executed prefix, and the remaining actions with observation history.

To ground this motion proposal in visual evidence, we map predicted command poses into the camera frame via calibrated reprojection, comparing them against observed object locations, historical pixel motion, and RGB-D back-projections. Combined with proprioceptive state, this visual-geometric correspondence allows the agent to distinguish a displaced approach target from a low-level execution slip or premature grasp.

Based on this diagnosis, the agent selects between two lightweight recovery routes. When the interaction target itself is inaccurate, the agent revises $G_k$ and invokes the guided VLA again to synthesize a new globally coherent chunk. When the generated trajectory already captures the right maneuver structure, the agent reuses the proposal and applies local adjustments via robot-control tools, such as shifting an approach offset, lift height, or gripper release timing. Together with early rejection of low-confidence targets ($C_t<e^{-1}$), this closes the bidirectional loop: agent-specified geometry steers VLA generation, and the VLA's motion proposal guides the agent's next correction.

\section{Experiments}
\label{sec:experiments}
\begin{table*}[!t]
\centering\footnotesize\fontfamily{lmr}\selectfont
\caption{\textbf{VLA tool use by agents on LIBERO-Plus (\%).} We compare standalone MR-0 (with training-data GT or offline Sol targets) and online Sol + MR-0 systems. Shaded rows use InteractVLA. Bold and underline denote best and second-best results. $^{*}$Retains base behavior under robot-init perturbations.}
\label{tab:libero}
\renewcommand{\arraystretch}{1.15}
\setlength{\tabcolsep}{5pt}
\resizebox{\linewidth}{!}{%
\begin{tabular}{l | c c c c c c c || c}
\toprule
\textbf{Method} & \textbf{Background} & \textbf{Light} & \textbf{Language} & \textbf{Noise} & \textbf{Layout} & \textbf{Camera} & \textbf{Robot Init.} & \textbf{Overall} \\
\midrule
MR-0 & 99.8 & 95.8 & 90.4 & 90.3 & 96.9 & 85.1 & 64.0 & 88.5 \\
\rowcolor{blue!4}
\quad + InteractVLA (Train-data) & 99.8 & \second{99.8} & \second{97.2} & 95.2 & 97.9 & 93.2 & 64.0$^{*}$ & 91.3 \\
\rowcolor{blue!4}
\quad + InteractVLA (Offline Sol) & \second{99.9} & \best{100.0} & 94.3 & 94.4 & \second{98.2} & \second{95.9} & 81.0 & \second{94.5} \\
\midrule
GPT-6.1 Sol + MR-0 & \best{100.0} & 94.7 & 91.6 & \best{96.3} & 89.5 & 87.6 & \second{94.3} & 93.1 \\
\rowcolor{blue!4}
\quad + InteractVLA & \second{99.9} & \best{100.0} & \best{100.0} & \second{95.9} & \best{98.4} & \best{98.4} & \best{96.2} & \best{98.3} \\
\bottomrule
\end{tabular}%
}
\end{table*}

\begin{table}[!t]
\centering\footnotesize\fontfamily{lmr}\selectfont
\caption{\textbf{VLA tool use by agents on RoboDojo.} Success rates (\%). Train-Data Guidance uses reference poses from training data; InteractVLA adds our online test-time guidance. Sol denotes GPT-6.1 Sol.}
\label{tab:dojo}
\renewcommand{\arraystretch}{1.15}
\setlength{\tabcolsep}{5pt}
\resizebox{\linewidth}{!}{%
\begin{tabular}{l | c c | c c}
\toprule
& \multicolumn{2}{c|}{\textbf{VLA}} & \multicolumn{2}{c}{\textbf{Sol + VLA}} \\
\cmidrule(lr){2-3}\cmidrule(lr){4-5}
\textbf{Dimension} & \textbf{Baseline} & \textbf{InteractVLA} & \textbf{Baseline} & \textbf{InteractVLA} \\
\midrule
Generalization & 8.2 & \best{28.0} & 21.6 & \second{26.0} \\
Precision & 7.8 & 4.0 & \best{9.7} & \second{8.3} \\
Long-Horizon & \best{18.6} & \second{15.3} & 8.9 & 12.5 \\
Memory & 8.9 & \best{24.0} & 16.5 & \second{22.2} \\
Open & 2.4 & 3.2 & \second{31.2} & \best{39.6} \\
\midrule
\textbf{Overall} & 9.2 & 12.17 & \second{18.0} & \best{26.0} \\
\bottomrule
\end{tabular}%
}
\end{table}

\begin{figure}
  \centering
  \includegraphics[width=0.47\textwidth]{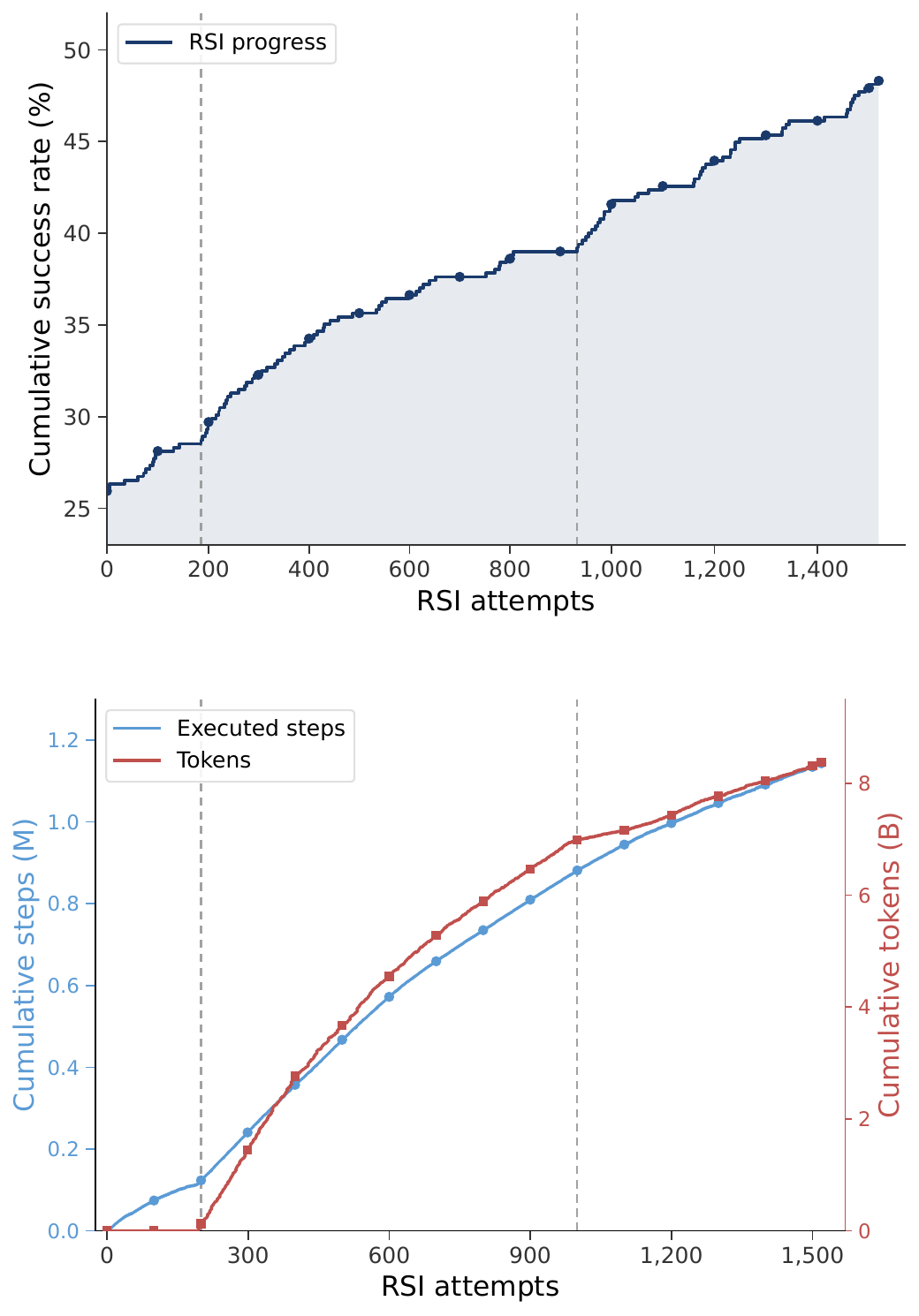}
  \caption{\textbf{RSI improves task success through exploration and experience consolidation on RoboDojo.} \emph{Top:} Cumulative success rate on RoboDojo over successive RSI attempts. \emph{Bottom:} Cumulative simulation steps and token consumption. Initial VLA attempts incur no agent-token cost. The agent then specifies geometric targets to explore alternative motions and discover new solutions through InteractVLA. As experience accumulates, the agent shifts toward consolidating existing strategies for task completion, slowing the growth of both execution steps and token consumption and increasing success rate substantially.}
  \label{fig:rsi}
\end{figure}
\begin{figure*}
  \centering
  \includegraphics[width=\textwidth]{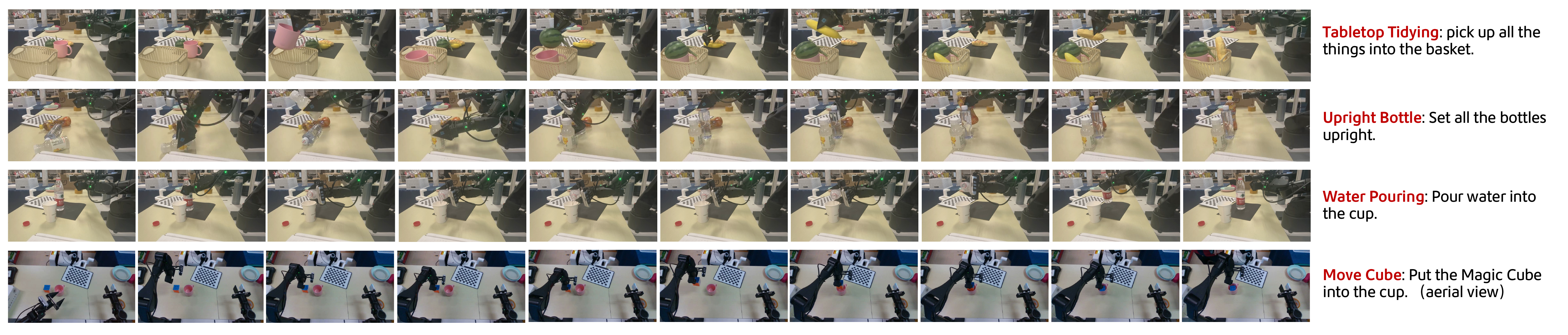}
  \caption{\textbf{InteractVLA enables diverse real-world manipulation through interactive VLA control.} From top to bottom, the sequences show Tabletop Tidying, Upright Bottle, Water Pouring, and Move Cube on the Dexmal platform. Each row presents successive stages of task execution from left to right, demonstrating coordinated grasping, reorientation, and placement under agent guidance. For Move Cube, we show the robot's overhead camera view to highlight the tight clearance between the cube and the cup, illustrating the precision required for placement. Full video demonstrations are available at \url{https://lingwupro.github.io/InteractVLA/}.}
  \label{fig:real}
\end{figure*}
\begin{figure*}
  \centering
  \includegraphics[width=\textwidth]{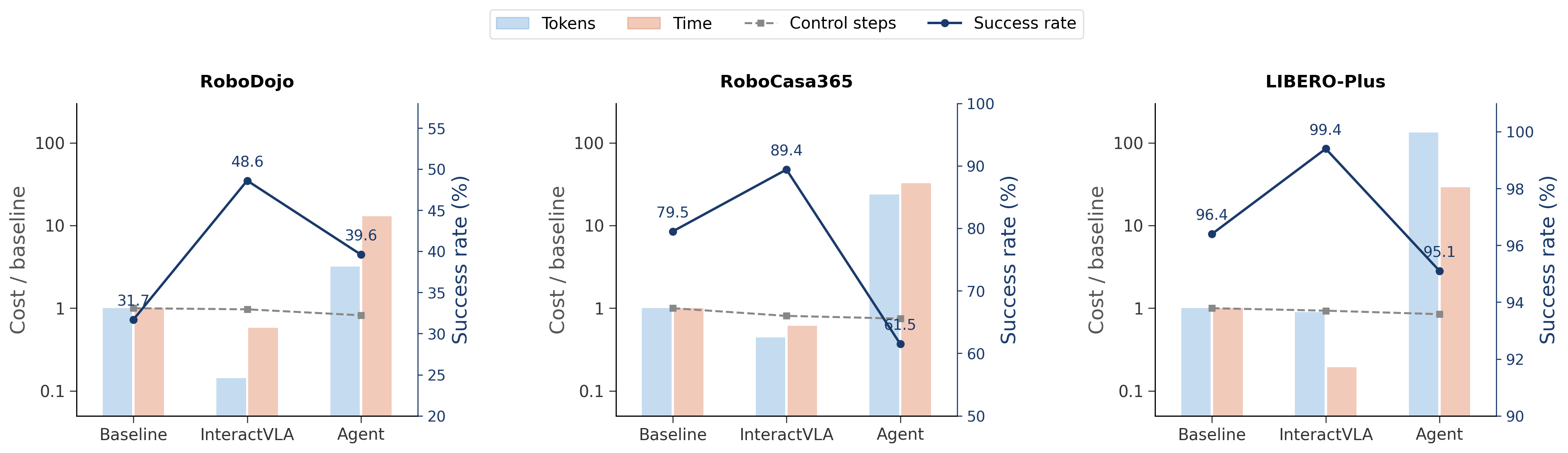}
  \caption{\textbf{InteractVLA improves task success with lower cumulative RSI cost.} We compare Agent + VLA, Agent + InteractVLA, and Agent Only on three benchmarks. Bars show agent API tokens and end-to-end time, and the dashed line shows environment control steps, all accumulated over every RSI attempt per case and averaged across cases. Costs are normalized to Agent + VLA on the logarithmic left axis, while the solid line reports the corresponding final success rates on the right.}
  \label{fig:efficiency}
\end{figure*}
\begin{table*}[!t]
\centering\footnotesize\fontfamily{lmr}\selectfont
\caption{\textbf{RSI comparison on LIBERO-Plus (\%).} Standalone VLA results are from~\cite{liberoplus}; agent baselines marked $^{\dagger}$ are from RV-ICL~\cite{rvicl} on the stratified 120-variant subset using GPT-6 Astra. All methods follow their official evaluation protocols.}
\label{tab:libero_rsi}
\renewcommand{\arraystretch}{1.15}
\setlength{\tabcolsep}{5pt}
\resizebox{\linewidth}{!}{%
\begin{tabular}{l c | c c c c c c c || c}
\toprule
\textbf{Method} & \textbf{Planner} & \textbf{Background} & \textbf{Light} & \textbf{Language} & \textbf{Noise} & \textbf{Layout} & \textbf{Camera} & \textbf{Robot Init.} & \textbf{Overall} \\
\midrule
OpenVLA-OFT & -- & \second{93.3} & 88.7 & 79.5 & 75.8 & 74.2 & 56.4 & 31.9 & 69.6 \\
$\pi_0$ & -- & 81.4 & 85.0 & 58.8 & 79.0 & 68.9 & 13.8 & 6.0 & 53.6 \\
$\pi_0$-FAST & -- & 73.2 & 73.2 & 61.0 & 74.4 & 68.8 & 65.1 & 21.6 & 61.6 \\
RIPT-VLA & -- & 91.6 & 88.4 & 77.6 & 73.5 & 74.2 & 55.2 & 31.2 & 68.4 \\
\midrule
HarnessVLA$^{\dagger}$ & GPT-6 Astra & 88.2 & 76.5 & \best{100.0} & 94.1 & 82.4 & 83.3 & 82.4 & 86.7 \\
RV-ICL$^{\dagger}$ & GPT-6 Astra & \best{100.0} & \second{94.1} & \best{100.0} & \best{100.0} & \second{88.2} & \second{94.4} & \second{94.1} & \second{95.8} \\
\midrule
\rowcolor{blue!4}
InteractVLA + RSI & GPT-6.1 Sol & \best{100.0} & \best{100.0} & 99.9 & \second{98.4} & \best{99.2} & \best{99.5} & \best{98.9} & \best{99.4} \\
\bottomrule
\end{tabular}%
}
\end{table*}

\begin{table*}[!t]
\centering\footnotesize\fontfamily{lmr}\selectfont
\caption{\textbf{RSI comparison on RoboDojo (\%).} We compare published robot agents~\cite{robodojo_lb} with our Astra-based InteractVLA + RSI system under official leaderboard protocols.}
\label{tab:dojo_rsi}
\renewcommand{\arraystretch}{1.15}
\setlength{\tabcolsep}{5pt}
\resizebox{\linewidth}{!}{%
\begin{tabular}{ll | c c c c c || c}
\toprule
\textbf{Method} & \textbf{Type} & \textbf{Generalization} & \textbf{Precision} & \textbf{Long-Horizon} & \textbf{Memory} & \textbf{Open} & \textbf{Overall} \\
\midrule
PhysicalRSI & Agent+VLA & 15.62 & \second{32.50} & \second{37.33} & 46.56 & 24.92 & 31.38 \\
WorldScape-R$^2$SI & Agent+WAM & \second{35.83} & 15.50 & 34.25 & 63.67 & 47.50 & 39.35 \\
EmbodiedEvo & Agent & 30.83 & 16.25 & 21.75 & \second{76.67} & \second{49.25} & 38.95 \\
HexaShell-Preview & Agent & 34.50 & 32.00 & 31.50 & \best{81.00} & \best{55.25} & \second{46.85} \\
GPT-6 Astra & Agent & 30.50 & 4.00 & 8.25 & 38.67 & 31.00 & 22.48 \\
\midrule
\rowcolor{blue!4}
InteractVLA + RSI & Agent+VLA & \best{52.78} & \best{39.58} & \best{45.83} & 63.89 & 41.67 & \best{48.62} \\
\bottomrule
\end{tabular}%
}
\end{table*}

\begin{table}[!t]
\centering\footnotesize\fontfamily{lmr}\selectfont
\caption{\textbf{RSI comparison on RoboCasa365 (\%).} Published VLA and agent baselines are from~\cite{robocasa_lb,harness}. $^{\dagger}$HexaShell 1.5 uses the pretrain50-kitchen protocol with task memory~\cite{hexa_casa_submission}.}
\label{tab:casa_rsi}
\renewcommand{\arraystretch}{1.15}
\setlength{\tabcolsep}{5pt}
\resizebox{\linewidth}{!}{%
\begin{tabular}{l | c c c || c}
\toprule
\textbf{Method} & \textbf{Atomic-Seen} & \textbf{Composite-Seen} & \textbf{Composite-Unseen} & \textbf{Overall} \\
\midrule
Meituan-Robotics-0 & 77.3 & 57.8 & 27.1 & 55.0 \\
HarnessVLA~\cite{harness} & 87.8 & 43.8 & 42.5 & 59.2 \\
HexaShell 1.5$^{\dagger}$ & \second{91.1} & \second{85.0} & \second{80.0} & \second{85.6} \\
\midrule
\rowcolor{blue!4}
InteractVLA + RSI & \best{93.6} & \best{89.2} & \best{84.8} & \best{89.4} \\
\bottomrule
\end{tabular}%
}
\end{table}

\begin{table}[!t]
\centering\footnotesize\fontfamily{lmr}\selectfont
\caption{\textbf{Real-world manipulation on Dexmal.} Task score and execution time (s) across four tasks.}
\label{tab:real_world}
\renewcommand{\arraystretch}{1.15}
\setlength{\tabcolsep}{5pt}
\resizebox{\linewidth}{!}{%
\begin{tabular}{l | c c | c c | c c}
\toprule
& \multicolumn{2}{c|}{\textbf{VLA}}
& \multicolumn{2}{c|}{\textbf{Agent + VLA}}
& \multicolumn{2}{c}{\textbf{Agent + InteractVLA}} \\
\cmidrule(lr){2-3}\cmidrule(lr){4-5}\cmidrule(lr){6-7}
\textbf{Task}
& \textbf{Score} & \textbf{Time (s)}
& \textbf{Score} & \textbf{Time (s)}
& \textbf{Score} & \textbf{Time (s)} \\
\midrule
Tabletop Tidying & 31.2 & 97.4 & 65.5 & 372.0 & \textbf{92.5} & 295.8 \\
Upright Bottle  & 16.9 & 63.8 & 37.9 & 226.8 & \textbf{74.6} & 217.1 \\
Water Pouring   & 41.6 & 27.4 & 38.5 & 97.0  & \textbf{79.2} & 71.2 \\
Move Cube      & 51.5 & 22.9 & 86.7 & 165.2 & \textbf{97.6} & 67.5 \\
\bottomrule
\end{tabular}%
}
\end{table}
\begin{table*}[!t]
\centering
\renewcommand{\arraystretch}{1.15}
\begin{minipage}[t]{0.49\linewidth}
\centering\footnotesize\fontfamily{lmr}\selectfont
\caption{\textbf{Ablation on target-time alignment (\%).} All variants share the same VLA backbone, pose target, and RSI budget.}
\label{tab:time_ablation}
\setlength{\tabcolsep}{5pt}
\resizebox{\linewidth}{!}{%
\begin{tabular}{l | c c c}
\toprule
\textbf{Alignment} & \textbf{LIBERO-Plus} & \textbf{RoboDojo} & \textbf{RoboCasa365} \\
\midrule
Fixed final step & 96.8 & 39.1 & 83.0 \\
Uniform over steps & 95.6 & 34.7 & 79.8 \\
Hard minimum & 98.4 & 44.2 & 86.1 \\
\midrule
\rowcolor{blue!4}
Soft-time (InteractVLA) & \best{99.4} & \best{48.6} & \best{89.4} \\
\bottomrule
\end{tabular}%
}
\end{minipage}\hfill
\begin{minipage}[t]{0.49\linewidth}
\centering\footnotesize\fontfamily{lmr}\selectfont
\caption{\textbf{Ablation on action-informed feedback (\%).} All variants use soft-time guidance under an identical RSI budget.}
\label{tab:feedback_ablation}
\setlength{\tabcolsep}{5pt}
\resizebox{\linewidth}{!}{%
\begin{tabular}{l | c c c}
\toprule
\textbf{Feedback} & \textbf{LIBERO-Plus} & \textbf{RoboDojo} & \textbf{RoboCasa365} \\
\midrule
Observations only & 97.9 & 40.3 & 84.1 \\
+ Generated actions & 98.7 & 44.8 & 87.0 \\
\midrule
\rowcolor{blue!4}
+ Reprojection (InteractVLA) & \best{99.4} & \best{48.6} & \best{89.4} \\
\bottomrule
\end{tabular}%
}
\end{minipage}
\vspace{-20px}
\end{table*}

We evaluate InteractVLA across four parts: (1) how our bidirectional geometric interface transforms online VLA tool use by agents, (2) how controllable VLA execution unlocks recursive self-improvement (RSI), (3) real-world bimanual manipulation, and (4) ablations on inference efficiency, temporal alignment, and feedback design.

\paragraph{Setup.}
We use MR-0~\cite{team2026meituan} as the pretrained VLA backbone on LIBERO-Plus~\cite{liberoplus}, RoboDojo~\cite{robodojo}, and RoboCasa365~\cite{robocasa365}. Because GPT-6 Astra offers stronger reasoning at a higher API cost than GPT-6.1 Sol (abbreviated as Sol), we use Astra for RSI on RoboDojo and RoboCasa365 while adopting the more cost-efficient Sol for online tool-use and LIBERO-Plus evaluations to balance rollout budgets~\cite{rasheed2026hard,li2026agentic}; substituting Astra in those Sol-based settings would theoretically yield even stronger performance. We compare three guidance modes: GT guidance using training-data reference poses, offline guidance reusing fixed agent-predicted targets, and online guidance where the agent iteratively updates targets and refines motions from execution feedback.

\subsection{VLA Tool Use by Agents}
\label{sec:agent_tool_use}
We first examine whether upgrading a VLA from a language-only black box to a geometrically steerable tool improves online agentic manipulation across perturbations.

\paragraph{LIBERO-Plus.}
As shown in Table~\ref{tab:libero}, steering standalone VLA with explicit pose targets immediately unlocks latent execution capability even without online replanning, raising overall success from 88.5\% to 91.3\% with GT targets and 94.5\% with offline Sol targets. In the online agent-VLA loop, replacing language-only invocation (Sol + VLA, 93.1\%) with InteractVLA boosts overall success to 98.3\%. The gains are most pronounced under spatial shifts where language alone cannot ground the shifted geometry, jumping from 89.5\% to 98.4\% under layout perturbations and from 87.6\% to 98.4\% under camera viewpoint shifts. Explicit pose targets let the agent anchor the target location in the current scene while relying on the VLA to synthesize the intervening motion.

\paragraph{RoboDojo.}
Table~\ref{tab:dojo} extends this analysis to challenging bimanual manipulation. While static GT pose targets lift standalone VLA from 9.2\% to 12.17\% overall through gains in generalization (28.0\%) and memory (24.0\%), they slightly degrade precision and long-horizon tasks, indicating that a static spatial constraint alone cannot resolve complex multi-step sequencing or contact dynamics. Closing the online loop with InteractVLA overcomes this bottleneck, lifting Sol + VLA from 18.0\% to 26.0\% overall and yielding consistent improvements across generalization (26.0\%), long-horizon (12.5\%), memory (22.2\%), and open-ended manipulation (39.6\%). By inspecting motion proposals and updating targets as execution unfolds, the agent dynamically adapts its geometric intent rather than committing to a fixed target stream. Geometric guidance and online refinement make pretrained VLAs more effective tools for agentic manipulation, which is also useful for \emph{Recursive Self-Improvement (RSI)}.

\subsection{Performance on Recursive Self-Improvement}
\label{sec:rsi}
During RSI, the agent explores in practice environments sharing the target task semantics but initialized with distinct random seeds and scene layouts, while the actual evaluation scenes remain strictly unexposed. Because InteractVLA provides explicit target inputs and interpretable trajectory outputs, the agent can diagnose failures across practice rollouts, consolidate reusable strategies (Figure~\ref{fig:rsi}), and transfer them to held-out evaluation instances.

\paragraph{LIBERO-Plus.}
In Table~\ref{tab:libero_rsi}, InteractVLA + RSI achieves a state-of-the-art 99.4\% overall success rate, reaching 100\% on background and lighting perturbations and at least 98.4\% across all seven categories. It substantially outperforms standalone finetuned VLAs (OpenVLA-OFT at 69.6\%, RIPT-VLA at 68.4\%) and prior agentic VLA systems including HarnessVLA (86.7\%) and RV-ICL (95.8\%)~\cite{rvicl}.

\paragraph{RoboDojo.}
As shown in Table~\ref{tab:dojo_rsi}, InteractVLA + RSI with Astra reaches 48.62\% overall success, outperforming the closest Agent+VLA baseline PhysicalRSI (31.38\%)~\cite{physicalrsi}, the Agent+WAM method WorldScape-R$^2$SI (39.35\%)~\cite{worldscapersi}, and pure agent controllers EmbodiedEvo (38.95\%) and HexaShell-Preview (46.85\%)~\cite{robodojo_lb}. Notably, InteractVLA leads across generalization (52.78\%), precision (39.58\%), and long-horizon execution (45.83\%), confirming that steering a motor policy with learned geometric strategies handles fine-grained contact and multi-stage tasks far better than raw agent control or unsteered VLA calls~\cite{zhang2026unexpected}.

\paragraph{RoboCasa365.}
On large-scale household manipulation (Table~\ref{tab:casa_rsi}), InteractVLA + RSI achieves 89.4\% overall success (93.6\% Atomic-Seen, 89.2\% Composite-Seen, and 84.8\% Composite-Unseen), outperforming the base MR-0 policy (55.0\%), HarnessVLA (59.2\%), and HexaShell 1.5 (85.6\%)~\cite{hexa_casa_submission,robocasa_lb,harness}. The 84.8\% success rate on Composite-Unseen highlights how geometric steering and action-informed refinement generalize atomic skills to unseen multi-stage task compositions.

\subsection{Real-World Experiments}
\label{sec:real_world}

\paragraph{Setup.}
We deploy InteractVLA on a physical Dexmal platform across four tasks (\emph{Tabletop Tidying}, \emph{Upright Bottle}, \emph{Water Pouring}, and \emph{Move Cube}; Figure~\ref{fig:real}). We collect 50 demonstrations per task to train an identical MR-0 backbone and evaluate each method over 30 trials per task, comparing standalone VLA, language-only Agent + VLA, and Agent + InteractVLA across mean task score, execution time, and agent API calls.

\paragraph{Results.}
Across all four real-world tasks (Table~\ref{tab:real_world}), Agent + InteractVLA simultaneously achieves the highest task success and the lowest latency and API overhead compared to language-only Agent + VLA. On Tabletop Tidying and Upright Bottle, InteractVLA boosts score from 65.5 to 92.5 and from 37.9 to 74.6 while cutting API calls by over half on tidying (23.5 to 9.9). The advantage of geometric guidance is particularly striking on precision-critical tasks: on Water Pouring, language-only retries stagnate at 38.5, whereas InteractVLA more than doubles performance to 79.2; on Move Cube, InteractVLA reaches 97.6 points while reducing execution time by 59\% (165.2s to 67.5s) and API calls by 67\% (14.4 to 4.7). Rather than repeatedly resetting the scene after blind retries, the agent steers the policy to the exact interaction pose in a few calls.

\subsection{Ablation Studies}
\label{sec:ablations}

\paragraph{Inference efficiency.}
Figure~\ref{fig:efficiency} compares Agent + VLA, which invokes the policy through language instructions; Agent + InteractVLA, which adds geometric guidance and action-informed refinement; and Agent only, which directly generates low-level commands. Costs accumulate all RSI attempts, including failures and retries, for each case and are then averaged across cases. End-to-end time includes agent API latency, VLA inference and guidance computation, and environment execution. Across all three benchmarks, Agent + InteractVLA achieves higher success with lower token consumption and total execution time than both alternatives.

\paragraph{Target-time alignment.}
Table~\ref{tab:time_ablation} isolates the impact of our temporal alignment objective under an identical VLA backbone, pose target, and RSI budget. Averaging pose error uniformly across all steps performs worst (34.7\% on RoboDojo) because it pulls the entire approach trajectory toward the endpoint, while constraining only the fixed final step (39.1\%) imposes rigid timing. Selecting the hard minimum-error step improves flexibility (44.2\%), and our soft-time alignment achieves the best performance across all benchmarks (99.4\%, 48.6\%, and 89.4\%), confirming the benefit of smoothly sharpening temporal assignment during denoising at test time.

\paragraph{Action-informed feedback.}
Table~\ref{tab:feedback_ablation} evaluates the reverse interface by progressively enriching the feedback returned to the agent. Starting from visual execution observations alone (40.3\% on RoboDojo, 84.1\% on RoboCasa365), exposing the VLA's generated action chunks raises success to 44.8\% and 87.0\% by revealing the policy's intended continuation. Adding calibrated camera reprojection further boosts performance to 48.6\% and 89.4\%, proving that aligning 3D action proposals with 2D visual evidence is crucial for accurate spatial diagnosis and local motion refinement.
\section{Related Work}
\paragraph{Classifier-free and geometric guidance.}
Classifier-free guidance (CFG) steers generative sampling by contrasting conditional and unconditional score estimates~\cite{ho_cfg,cai2026tau_0,xu2026representation}, as adapted to goal-conditioned robot policies in BESO~\cite{beso}. Manipulation objectives can also be specified via 3D value maps~\cite{voxposer}, relational keypoints~\cite{rekep}, 3D trajectories~\cite{hamster,hamster3d}, or VLM-written reward functions in VLS~\cite{vls}. However, VLS relies on heuristic VLM-synthesized trajectory rewards with costly multi-particle resampling and does not resolve latent arrival timing for complete 6-DoF poses. In contrast, InteractVLA differentiates sparse agent-specified poses through the action expert within a single flow trajectory via soft-time target alignment, jointly resolving the approach motion and arrival step.

\paragraph{Agentic manipulation and runtime correction.}
Recent frameworks pair high-level agents with pretrained VLAs via retry mechanisms~\cite{harness}, recursive video in-context learning~\cite{rvicl}, spatial affordance intervention~\cite{afi}, or phase-specific runtime action offsets~\cite{actionground}. Meanwhile, recent test-time flow steering methods adapt frozen policies using human interventions~\cite{tang2026preferenceflow} or execution traces~\cite{zhang2026traceflow}, yet lack a closed-loop interface for autonomous agentic reasoning. While these works motivate inspecting and correcting policy behavior rather than treating VLA calls as atomic, they either rely on external primitives for recovery or apply black-box residual offsets without steering internal generation~\cite{liufu2026repo,yancan}. InteractVLA establishes a closed-loop bidirectional geometric interface: it steers native flow generation toward explicit 6-DoF targets in the forward pass and reprojects generated action chunks into the camera frame in the reverse pass, enabling the agent to choose between guided regeneration and local trajectory refinement.
\vspace{-3px}
\section{Conclusion}
\vspace{-3px}
We presented InteractVLA, a test-time framework that transforms pretrained flow-based VLAs from opaque language-only primitives into geometrically steerable and inspectable tools for vision-language agents. By coupling differentiable test-time geometric guidance and soft-time target alignment in the forward direction with action-informed agentic refinement in the reverse direction, InteractVLA enables agents to precisely steer policy generation, diagnose spatial errors via calibrated reprojections, and iteratively consolidate manipulation strategies through recursive self-improvement. Extensive benchmark and real-world evaluations demonstrate that this bidirectional geometric interface unlocks substantial gains in both manipulation success and inference efficiency.

\section*{Acknowledgements}
This work is supported by the National Key Research \& Development Plan (2023YFF0725100) and the National Natural Science Foundation of China (92570121, 62322214, U23A20299, U24B20144). We thank Xiaofu Chen, Han Pan and Jinghong Liu for real-world experiments and Dongting Li for discussion.
{
    \small
    \bibliographystyle{ieeenat_fullname}
    \bibliography{main}
}


\end{document}